\documentclass[twoside,leqno,twocolumn]{article}

\usepackage[letterpaper]{geometry}

\usepackage{ltexpprt}
\usepackage{hyperref}
\usepackage{amsmath,amssymb,amsfonts}
\usepackage{graphicx}
\usepackage{textcomp}
\usepackage{xcolor}
\usepackage{algorithm,algcompatible}

\usepackage{subfigure}

\newtheorem{definition}{Definition}

\begin{document}

\newcommand\relatedversion{}
\renewcommand\relatedversion{\thanks{The full version of the paper can be accessed at \protect\url{https://arxiv.org/abs/1902.09310}}} 

\title{\Large 
PMP: Privacy-Aware Matrix Profile against Sensitive Pattern Inference \\
for Time Series}

\author{Li Zhang\thanks{
George Mason University, \{lzhang18, jessica\}@gmu.edu}\qquad 
Jiahao Ding\thanks{
University of Houston, jding7@uh.edu}\qquad
Yifeng Gao\thanks{
University of Texas Rio Grande Valley, yifeng.gao@utrgv.edu}\qquad Jessica Lin\footnotemark[1]}

\date{}

\maketitle


\fancyfoot[R]{\scriptsize{Copyright \textcopyright\ 2023 by SIAM\\
Unauthorized reproduction of this article is prohibited}}





\begin{abstract} \small\baselineskip=9pt 
Recent rapid development of sensor technology has allowed massive fine-grained time series data to be collected and set foundation for the development of data-driven services and applications. During the process, data sharing is often involved to allow the third-party modelers to perform specific time series data mining tasks based on the need of data owner. The high resolution of time series data brings new challenges in protecting privacy. On one hand, meaningful information in high-resolution time series shifts from concrete point values to local shape-based segments. On the other hand, numerous research have found that long shape-based patterns could contain more sensitive information and may potentially be extracted and misused by a malicious third party. However, the privacy issue for time series patterns is surprisingly seldom explored in privacy-preserving literature. In this work, we consider a new privacy preserving problem: preventing malicious inference on long shape-based patterns while preserving short segment information for the utility task performance. To mitigate the challenge, we investigate an alternative approach by sharing Matrix Profile (MP), which is a non-linear transformation of original data and a versatile data structure that supports many data mining tasks. We found that while MP can prevent the concrete shape leakage, the canonical correlation in MP index can still reveal the location of sensitive long pattern information. Based on this observation, we design two attacks named Location Attack and Entropy Attack to extract the pattern location from MP. To further protect MP from these two attacks, we propose a Privacy-Aware Matrix Profile (PMP) via perturbing the local correlation and breaking the canonical correlation in MP index vector. We evaluate our proposed PMP against baseline noise-adding methods through quantitative analysis and real-world case study to show the effectiveness of the proposed method. Our source code is available at \href{https://github.com/lzhang18/PMP}{https://github.com/lzhang18/PMP}. 

\end{abstract}

\section{Introduction}

The wide use of sensors in personal devices and other infrastructures have allowed the collection of high-resolution time series and boosted the demand for third-party data-oriented services and applications such as medical monitoring~\cite{yeh2016matrix}, industrial system prognostics~\cite{zhu2017tsc} and smart homes~\cite{yao2017deepsense}. For example, a farm owner (referred as \textbf{`data owner'} or \textbf{`owner'} for short) might collect massive data from their own smart sensor system to monitor the behavior of farm animals such as cows and chicken in real-time~\cite{abdoli2020fitbit}. Since the owner does not have the expertise to analyze the data, they might seek some third-party data mining service tasks (aka \textbf{`utility task'}) such as activity recognition (e.g., identify eating food or egg laying), or detecting animal sickness through anomaly detection. 

While this data sharing service brings benefits, there has been long-time concern for data sharing such as personal information leakage and data breach~\cite{avancha2012privacy}. Existing solution has been relying on Differential Privacy (DP)~\cite{fan2013adaptive, xiao2015protecting} to hide the concrete data values associated with sensitive information. However, the high resolution of time series data brings new challenges in privacy protection. The concrete data values become less important as meaningful information shifts to the shape of small subsequences~\cite{esling2012time}. On the other hand, numerous research  \cite{gao2018exploring, yeh2016matrix, zhumatrix} have found that long shape-based patterns could contain more sensitive information and may potentially be extracted and misused by a malicious third party. For example, a malicious modeler might detect a day-long pattern that infer farmer's daily work route even though such day-long patterns is unrelated to any utility task mentioned above, which typically relies on minute or hour-long time series segments~\cite{erdemir2020privacy, tyagi2020survey}. However, the privacy issue for time series patterns is surprisingly seldom explored in privacy-preserving literature.

\begin{figure}[t] \centering
 \subfigure[Existing perturbed data sharing pipeline\label{Traditional1}]
 {\includegraphics[width=80mm]{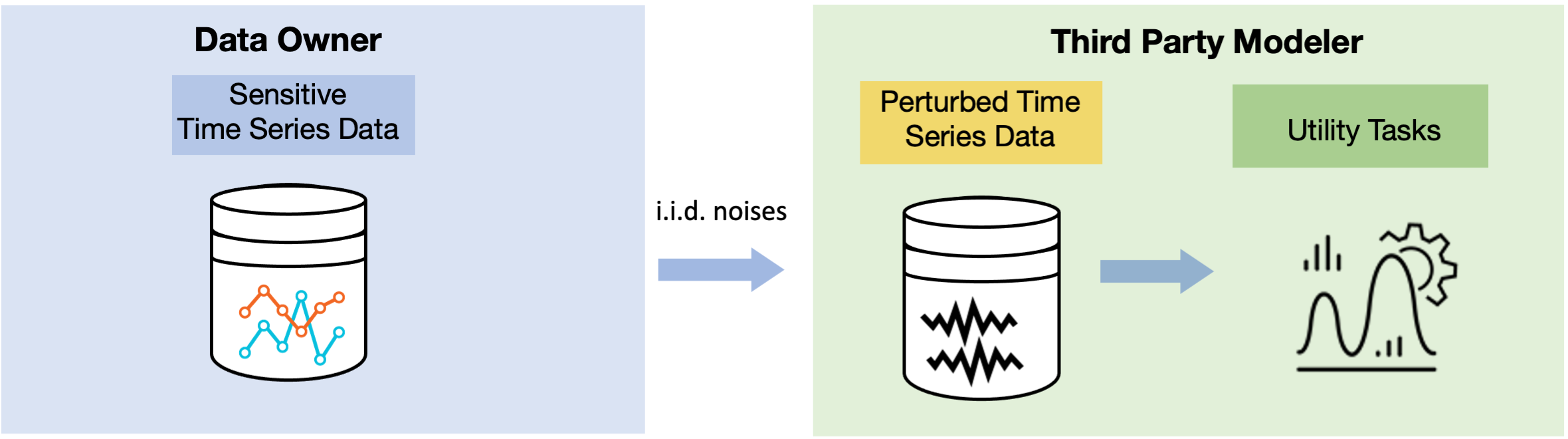}}\vspace{-0.1in}
\subfigure[ Proposed Private Matrix Profile (PMP) Data Sharing and Mining pipeline. \label{MP_sharing2}]
  { \includegraphics[width=84mm]{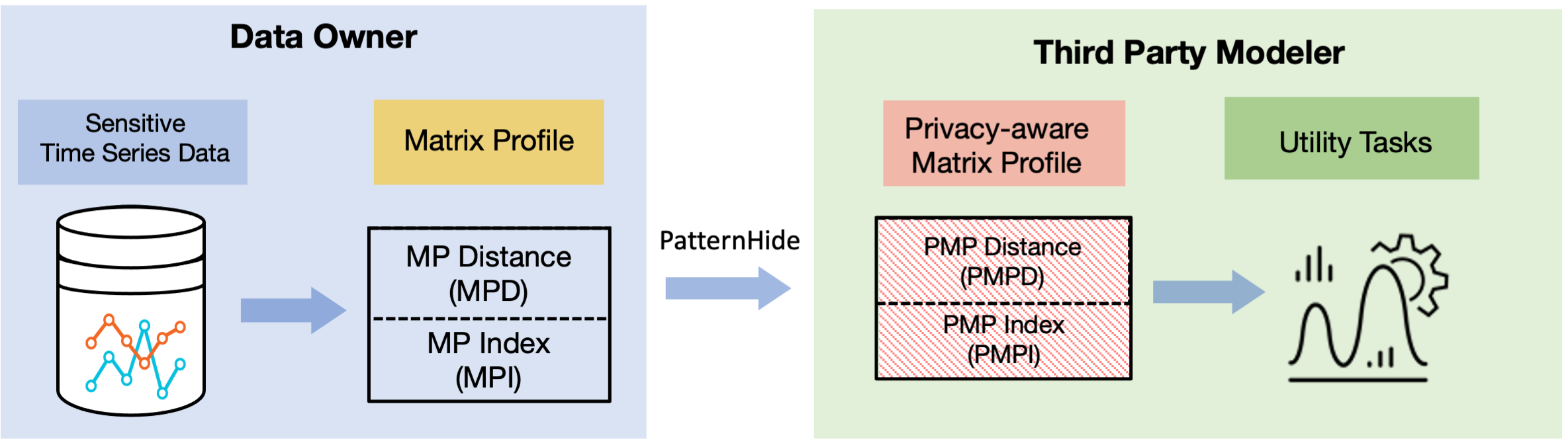}} \vspace{-0.2in}
 \caption{Comparison between existing pipeline and our proposed pipeline. PMP is computed from the raw time series and then shared to the third party to protect the shape and location for long sensitive patterns while maintaining the utility on local patterns. } \label{fig:system1}\vspace{-2mm}
\end{figure}

In this work, we consider a new privacy preserving problem: preventing malicious inference on long shape-based patterns while preserving short segment information for the downstream task performance. 
As shown in Fig.~\ref{Traditional1}, most existing privacy-preserving approaches are based on differential privacy ~\cite{fan2013adaptive, xiao2015protecting} to sanitize the raw time series values with independent and identical distributed (i.i.d) noises and then share the perturbed data with the third party modeler. However, as pointed out by Xiao et al.~\cite{xiao2015protecting}, DP methods cannot protect the privacy of pattern, which consists of a set of contagious autocorrelated points. In fact, if we adopt previous DP methods to protect the long pattern, the amount of noises needed to perturb a long pattern region would be more than sufficient to disrupt local segments, essentially making the shared time series useless for the utility tasks (as will be illustrated in Section~\ref{adding_noise}).

To address the dilemma between utility of short segments and privacy (preventing malicious inference on long shape-based patterns), we investigate an alternative approach by sharing Matrix Profile (MP) instead. MP is recent proposed as an efficient and versatile data structure that records a single distance and the index of its closest match for each subsequence of a given length in time series. There are two advantages to using MP. First, MP is ideal for third-party modelers to use as a versatile intermediate feature to support most fundamental data mining tasks such as motif discovery, anomaly detection, and more complex tasks such as rule discovery, segmentation, and data summarizing. Second, we found that it is difficult for malicious third-party to recover the original time series and the patterns themselves by solely using Matrix Profile due to the use of Z-normalized Euclidean distance (as discussed in Section~\ref{sec: recover from z-norm}). However, we found that the canonical correlation in MP index can still reveal the location of long patterns. Based on the observation, we design two attacks named Location Attack and Entropy Attack to retrieve the pattern location from MP. To further protect MP from these two attacks, we propose a {Privacy-Aware Matrix Profile (PMP)} via perturbing the local correlation and breaking the canonical correlation in MP index vector. Our overall framework is shown in Figure 1. Instead of sharing the raw time series or perturbed time series, the third-party modeler would only have access to the PMP. The modeler could still perform the utility task(s) based on PMP, but they would not be able to infer the shapes nor the locations of long sensitive patterns. 


In summary, our contributions are listed as follows: 
\begin{itemize}
\item We consider a new privacy preserving problem: preventing malicious inference on long shape-based patterns while preserving short segment information for the downstream task performance. To the best of our knowledge, this is the first work to investigate this problem, and it cannot be protected by existing approaches. 
\vspace{-1mm}
\item We investigate an alternative approach by sharing Matrix Profile (MP), a non-linear transformation of original data and a versatile data structure supporting many data mining tasks. We found that MP can protect sensitive long time series pattern shape, but could still leak the pattern location due to correlation in MP Index (MPI).
\item We design two attacks based on location and entropy of MPI to extract sensitive pattern location from MP. 
\item We further propose a defense algorithm called $PatternHide$ to generate a Privacy-aware Matrix Profile (PMP), which can prevent the location leakage of sensitive patterns while keeping the downstream task performance.
\item We evaluate our proposed PMP against baseline methods through quantitative analysis and multiple real-world case studies to show the effectiveness of the proposed method.
\end{itemize}

\section{RELATED WORK}

In the last two decades, vast research efforts have been put into time series data mining tasks such as time series classification~\cite{bagnall2018uea}, discord discovery~\cite{keogh2005hot}, motif discovery~\cite{mueen2009exact}, and segmentation~\cite{gharghabi2017matrix}. Different from point-based tasks, pattern-based time series approaches are based on similarity on the subsequence level instead of point values to capture the notion of \textit{shape} information, which goes beyond point-values and reifies human natural understanding and visualization~\cite{esling2012time}. Pattern-based time series methods handle large scale time series well with excellent performance and interpretation. Recently, Matrix Profile (MP)~\cite{yeh2016matrix,zhumatrix} is proposed as an efficient and effective data representation on subsequence level and support most major fundamental tasks for downstream applications in a broad range of domains applications~\cite{yeh2016matrix,zhumatrix,zhu2020swiss}.  Existing work such as~\cite{zymbler2021matrix} typically use the raw data to compute MP as the first, feature generation step, and then design algorithms or models based on the computed MP. None of the work considers the use of MP in the context of data sharing with a third party, nor the privacy issues raised by sensitive patterns. 

Most existing approaches for the privacy-preserving release are designed for low-resolution time series based on differential privacy (DP) to protect the events where each event is associated with a single point. 
For example, Dwork et al.~\cite{dwork2010differential} proposed a binary tree based DP algorithm for single events in finite streams. Fan et al.~\cite{fan2013adaptive} presented FAST for realizing DP on user-based finite streams with a framework of sampling-and-filtering.
However, these approaches are designed for low-resolution time series and cannot be used to protect long sequence while maintaining the downstream task performance.
In addition, our problem is also different from PatternLDP~\cite{wang2020towards}, which is designed to protect point values from malicious attacks while preserving the utility of local patterns, whereas our problem is in the opposite direction to protect sensitive long pattern while keeping the utility of short patterns. 

In summary, these time series DP solutions cannot be directly applied to our setting, since these approaches mainly focus on computing the aggregated estimates (e.g., prefix-sum and moving average) of the values under DP, whereas our goal is to release subsequence-level representation that prevents leaking the information of long patterns while maintaining different downstream task performance.

Another line of research focuses on sharing encrypted time-series data. However, these methods can only support aggregation statistics computation and cannot support time series data mining tasks such as motif discovery and anomaly detection. Moreover, they mostly rely on oblivious RAM, secure multiparty computation, and function secret sharing \cite{burkhalter2020timecrypt, burkhalter2021zeph,dauterman2022waldo}, which require a significant number of cryptographic operations and secure communication channels.

To the best of our knowledge, there has been no existing work that offers solution for the data sharing issue or potential leakage of patterns from using Matrix Profile, nor the task to protect from long pattern inference while maintaining downstream task performance  on local segments. 
\vspace{-3mm}
\section{Background and Preliminaries}

In this section, we first review necessary time series related notations. \\
\textbf{Time Series} $T = [t_1, t_2, \ldots, t_n]$ is a set of observation ordered by time, where $t_i$ is a finite real number and $n$ is the length of time series $ T$.\\
\noindent \textbf{Subsequence} $S^T_{i,l}= [t_i, t_{i+1}, \ldots, t_{i+l-1}]$ of time series $T$ is a contiguous set of points starting from position $i$ with length $l$. Typically $l\ll n$, and $1\leq i\leq n-l+1$.\\

Previous work such as~\cite{yeh2016matrix,keogh2005hot} require subsequence comparison be non-trivial match, which prevents the comparison of subsequences that overlap more than 50\% of the length.

\noindent \textbf{Non-trivial Match} Given a time series $T$, a subsequence $S_{i,l}$ with length $l$ is considered a \textit{non-trivial match} of another subsequence $S_{j,l}$ of length $l$ if $|i-j|> l/2$.

In many applications, we are interested in finding similar “shapes” between two subsequences.  Z-normalized Euclidean distance is used to achieve scale and offset invariance~\cite{lin2007experiencing}.\\
\textbf{Z-normalized Euclidean Distance (Z-norm ED)}: $d(S_{p,l}, S_{q,l})$ of subsequences $S_{p,l}$, $S_{q,l}$ of length $l$ is computed as $\tiny {\sqrt{\sum_{m = 1}^{l}{(\frac{t_{p+m-1} - \mu_{p,l}}{\sigma_{p,l}} - \frac{t_{q+m-1} - \mu_{q,l}}{\sigma_{q,l}} })^2}}$,
where $\mu_{p,l}$, $\sigma_{p,l}$ and $\mu_{q,l}$, $\sigma_{q,l}$ are the means and standard deviations of subsequences $S_{p,l}$ and $S_{q,l}$, respectively. 

Z-normalization step is very critical, as noted in previous work — “without normalization time series similarity has essentially no meaning. More concretely, very small changes in offset rapidly dwarf any information about the shape of
the two time series in question"~\cite{keogh2003need}. There is additional benefit of preventing data leaking brought by utilizing Z-normalization distance, and we will discuss in details in Sec. 5.

\noindent \textbf{Distance Profile} Given a query subsequence $Q_l$ of length $l$ and a time series $T$, a distance profile $D_T(Q)$ is a vector containing the Z-normalized Euclidean distances between $Q$ and each subsequence of the same length in time series $T$. Formally, $D(Q_l, T) = [d(Q_l,S^T_1), d(Q_l,S^T_2), \cdots, d(Q_l,S^T_{n-l+1})]$. \\
\noindent \textbf{Matrix Profile Distance (MPD)}: Matrix Profile Distance of time series $T$ given subsequence length $l$ is a vector of the Z-normalized Euclidean distances between every subsequence $S_{i,l}$ and its nearest neighbor (most similar) subsequence in time series $T$. Formally, $MP = [\min(D(S_{1,l}, T)), \min(d(S_{2,l}, T)),\cdots, \min(d(S_{n-l+1,l},T))].$ \\
\noindent \textbf{Matrix Profile Index (MPI)}: Matrix Profile Index of time series $T$ is a vector of indices containing the index of the non-trivial match of nearest neighbor subsequence of subsequence $S_{i,l}$ in time series $T$. 
Formally, $\text{MPI} = [\arg\min(d(S_{1,l}, T), \cdots, \arg\min(d(S_{n-l+1,l},T)].$\\
Matrix Profile (MP) consists of two vectors, MPD and MPI. MP contains rich information about the data and the pattern location. For example, time series motif~\cite{mueen2009exact} can be found by exacting minimum value of MPD and the corresponding MPI. 

\section{Problem Statement}
In practice, companies may utilize sensitive data for data mining in order to provide better services. As shown in Fig. \ref{Traditional1}, some data owner (e.g., factories with smart sensors, e-commerce platforms and hospitals) may capture clients’ time series and send to cloud servers/third party modelers for data analysis. The third party modelers will run Matrix Profile and then design a model for improving the quality of data owners' service or production-related decision making. However, directly transmitting raw time series or perturbed time series to modelers would allow malicious third-party modelers to use the long patterns to infer sensitive information as we explained earlier in Introduction.

Since it is risky to share the data directly, a better protocol (Fig. \ref{MP_sharing2}) is to first generate the MPD and MPI with a given length for the utility tasks, and then send it to the external cloud servers/modelers for utility tasks. For convenience, this given length is referred as the \textbf{utility length} and denoted as $L_{util}$. There are two key research problems we should answer to comprehensively evaluate the privacy preserving performance of Matrix Profile:
 \begin{itemize}
    \item \textit{Does sharing MPD and MPI of length $L_{util}$ protect the shapes of long patterns?} 
    \item \textit{Does sharing MPD and MPI of length $L_{util}$ protect the locations for long patterns?}
\end{itemize}

\vspace{-2mm}
\section{Advantages of Matrix Profile for Privacy Protection}

To answer above questions, we conduct a comprehensive study of Matrix Profile in terms of the privacy protection for long patterns. \\
\noindent\textbf{1) Difficult to recover raw data from MPD and MPI}\label{sec: recover from z-norm} We found that it is very difficult to recover the concrete values of time series solely from the shared Matrix Profile because of Z-normalization. Recall that Z-normalization requires the knowledge of the mean and the standard deviation for both subsequences (for equation, see Z-normalized Euclidean Distance in Section~3). To extract entire the time series from MP, we would need to know the mean and standard deviation for every pair of subsequences. 
Thus, if we would like to recover the data from the distance, we have to solve for the values of data from pairwise distance between subsequences with the mean $\mu_{i,l}$ and standard deviation $\sigma_{i,l}$ as parameters in every equation. As $\sigma_i$ is a non-linear function of variable $t_i$ to $t_{i+l-1}$, it makes the inverse problem ill-defined. One could get infinitely many possible time series data as possible solutions satisfying a given MP, and hence cannot retrieve the original data. \\
\noindent \textbf{2) Difficulty to infer long pattern location from MPD} In addition, as pointed out by previous work \cite{hime}, ``the distance between a pair of short subsequences does not necessarily share similar behavior with the distance between long subsequences if the length difference is large''. As a result, MPD naturally suppress the location correlation between small pattern and the long pattern. Thus, it is difficult to recover large pattern location from a short length MPD. 

\vspace{-3mm}
\section{Threat Model and Attack Methods}

While MP provides these two advantages in protecting privacy, since both MPD and MPI have to be shared to fulfill task utility, we found that attacker still can potentially derive the sensitive pattern location from MPI. We first define the threat model as follows:

\textbf{Adversary Goal.} Given MPD and MPI, the adversary aims to locate a pair of sensitive motifs of a sensitive length and we assume it is much longer than utility length $L_{util}$. For convinience, we refer to this length as \textbf{attack length} and denote it as $L_{attack}$. Specifically, the adversary goal is to detect frequent patterns in time series, which is similar to motif discovery task~\cite{mueen2009exact,mueen2013enumeration,linardi2018matrix,mercer2021matrix} with one exception -- the attacker could only use the shared Matrix Profile of length $L_{util}$ to detect the long motif. 

Following the widely used evaluation criteria \cite{Gao2018,mueen2009exact,mueen2013enumeration,gao2018exploring}, we use \textit{the success rate} to measure how accurate the detected pair of patterns match the actual locations.

\textbf{Adversary Knowledge.} We assume a strong adversary (e.g., honest-but-curious modeler), who has no access to the sensitive time series data, but has the white-box access to the Matrix Profile Distance $\text{MPD}$ and Matrix Profile Index $\text{MPI}$ with short subsequence length $L_{util}$ pre-generated from the sensitive data.

\subsection{Attack Methods}

\begin{figure}
    \centering
    \includegraphics[width=80mm]{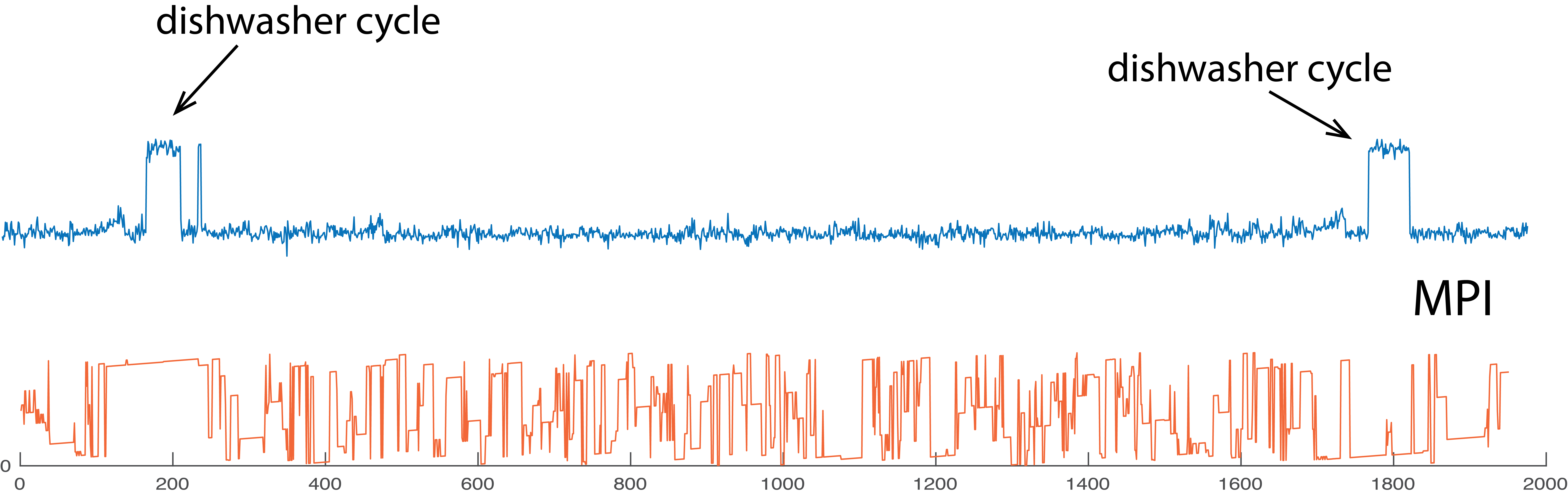}
    \vspace{-2mm}
    \caption{MPI index of a dishwasher time series with two dishwasher cycles}
    \label{fig:consec_subseq2}
\end{figure}

\begin{figure}
    \centering
    \includegraphics[width=80mm]{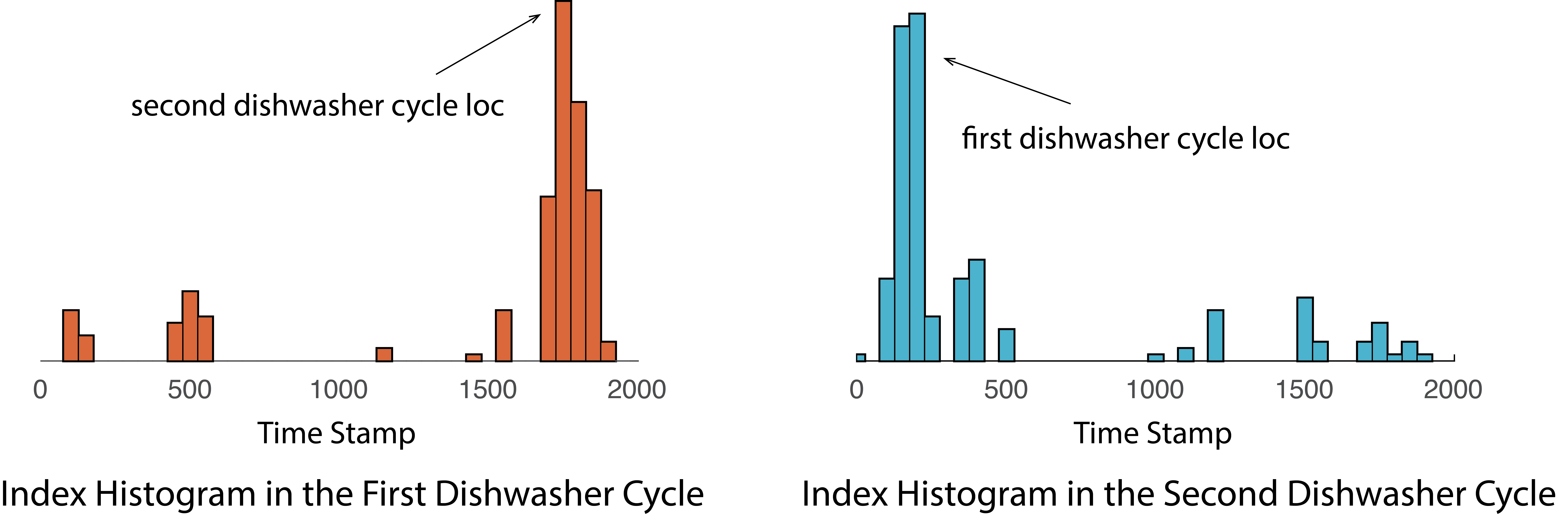}
    \vspace{-2mm}
    \caption{MPI index of a dishwasher time series with two dishwasher cycles}
    \label{fig:hist}
    \vspace{-2mm}
\end{figure}

We use a real-world time series to demonstrate how $\text{MPI}$ can leak long pattern information. Figure~\ref{fig:consec_subseq2}.top shows a snippet of a dishwasher power consumption time series. The time series contains two long dishwasher cycles. Obviously, at both pattern locations, the index evolution is more smooth and gradual than other region. This is because the consecutive 1-NN locations in the pattern region are likely similar since each subsequence is only offset by one point. Therefore, the characteristic of a $\text{MPI}$ block of length $L_{attack}$ may leak pattern locations. In addition, the histograms of the index among the two dish washer cycles are shown in Fig.~\ref{fig:hist}.left and Fig.~\ref{fig:hist}.right. The most frequent indices in both dishwasher cycles indeed correspond to the long pattern location of the other dishwasher cycle in the time series.




 \begin{algorithm}[t]\label{alg:attack1}
    \caption{Proposed Attack Algorithm}
 \begin{algorithmic}[1]
    \STATE \textbf{Input}: $\text{MPD}$, $\text{MPI}$, $L_{attack}$
    \STATE \textbf{Output}: long pair motif indices $idx1$, $idx2$ 
    \\{\color{blue} /* Calculate sum of distances in each sliding window of length $L_{attack}$*/}
    \STATE SumDist = Score($\text{MPD}$, $L_{attack}$)
    \STATE $idx1$ = $\arg\min$(SumDist)
    \\{\color{blue} /* Identify second motif index based 1st motif index*/}
    \STATE $idxPool$ = $\text{MPI}[idx1:idx1+L_{attack}-1]$
    \STATE $idx2$ = $\arg\max$ (GetIntervalFrequency($idxPool$))
    \STATE \textbf{return} $idx1$, $idx2$ 
  \end{algorithmic}
\end{algorithm}

Inspired by the above intuition, we introduce our basic framework of the attack method. The algorithm is illustrated in Algorithm 1. Given a sensitive pattern length $L_{attack}$, the attack algorithm will first scan through the MP time series with a sliding window equal to $L_{attack}$ and assigns the significant score (Line 4). The candidate with the most outstanding score will be identified as the location of the first pattern instance (i.e., $idx1$). Then, the second instance's location is identified by computing the histogram of all indexes belongs $\text{MPI}[idx1:idx1+L_{attack}-1]$. The centroid of the highest frequent bin is assigned as the second instance's location (e.g., the peak in Fig. \ref{fig:hist}) (Line 7-8). Different significant score in Line 4 will result in different attack methods. In this paper, we propose two scores based on $\text{MPI}$ for this framework:
\vspace{-2mm}
\begin{itemize}
    \item \textbf{Location-based Score:} The length of the consecutive index in the sliding window.
    \vspace{-1mm}
    \item \textbf{Entropy-based Score:} The negative entropy value given all indices in the sliding window.
\end{itemize}
We refer the first strategy as Location based Attack and the second strategy as Entropy-based Attack. Two attack methods could retrieve some degree of long pattern information if $\text{MP}$ is not protected.

Note that existing motif discovery algorithms~\cite{mueen2009exact,mueen2013enumeration,linardi2018matrix,mercer2021matrix} could not be used to detect the long motif of length $L_{attack}$ through the shared Matrix Profile of length $L_{util}$ due to the lack of access to the original time series data. However, our proposed attack methods can still find long motif from a single shared MPI. 

\vspace{-3mm}
\section{Defense Strategy}

In this section, we first introduce a new concept named Consecutive Index Block (CIB), and a new matrix profile, Masked Matrix Profile (Masked MP), which is highly related to the proposed algorithm. Finally, we introduce our proposed $PatternHide$ algorithm.
\vspace{-1mm}
\subsection{Consecutive Index Block and Masked Matrix Profile}

We introduce Consecutive Index Block (CIB) to capture the gradual changed regions in $\text{MPI}$ that highly related to long patterns. Specifically, \\
\noindent \textbf{Consecutive Index Block (CIB)} \label{cib}
Given a Matrix Profile index $\text{MPI}$, a Consecutive Index Block (CIB) $C_k$ consists of a set of consecutive index starting from index $k$ where for any index $i\in C_k$, we have $|\text{MPI}(i)- \text{MPI}(i+1)|< L_{attack}$.\\

Intuitively, the overall defense strategy of the proposed algorithm is perturbing and hiding any outstanding CIB regions in $\text{MPI}$ because long CIB potentially aligns with long pattern location and leads to pattern leakage. We next introduce Masked MP, the alternative ``fake'' MP used to hide real MP, while maintaining compatible functionality. \\
\noindent \textbf{Masked Matrix Profile} \label{mask_mp}
Given a mask vector $M$, Masked Matrix Profile is computed through the same process as Matrix Profile but exclude any masked locations in $M$ during computing $\text{MPI}$ and $\text{MPD}$.\\
\vspace{-5mm}
\subsection{Proposed Algorithm}
The proposed algorithm, \textit{PatternHide}, is described in Algorithm \ref{alg:defense}. Given a matrix profile $\text{MP}$, the algorithm consists of three steps. First, the algorithm obtains all the CIBs $\mathcal{C}$ in $T$ and forms a set of sensitive segments $\mathcal{S}$ that may leak the pattern information. A segment is sensitive if it is close (index difference is less than $L_{attack}$) to a long CIB $C_i$ with length greater than $L_{perm}$ (Line 4-8). Then the algorithm will perform the permutation step to generate ``fake'' 1-NN via masked matrix profile to break down long CIBs into small pieces that is similar to the non-sensitive area for every sensitive segment (Line 10-24). Finally, after checking all the sensitive segments, the algorithm will further examine existing cycles in MPI and modify any conflicting distance values to ensure consistency with the MPI.

\noindent \textbf{1)Breakdown Long CIBs in Sensitive Segments}  
The goal of this component is to modify the indices in any sensitive segments that potentially leak the long pattern, but still keep most of the functionality of the original matrix profile. To achieve this goal, we replace the MPD and MPI with Masked Matrix Profile (Line 10-23) which is computed with the sensitive area masked to ensure that no CIB of significantly large length exists. In the algorithm, if the consecutive index length is greater than a randomly generated threshold $L_{rnd}$ at time point $i$, we update the corresponding MPI and MPD with that of mask matrix profile. Specifically, given a sensitive segment $S$, the algorithm maintains a mask vector $M$ to control the permuted matrix profile (Line 10-11). Every time the condition in Line 14 is met, any subsequence overlapped with $\text{MPI}(i)$ is added into mask vector (Line 16) and triggers the permutation process (Line 17-20). In the process, the algorithm computes masked matrix profile with $M$ (Line 17) and replaces the remaining $\text{MPD}$ and $\text{MPI}$ in the sensitive segments with newly computed mask MP (Line 19-20). Then the algorithm samples another length threshold $L_{rnd}$ to prepare for the next pattern hiding operation (Line 13). The mask $M$ is set to zero vector after examining each sensitive segment. Note that the time complexity of this component is the same as computing a matrix profile because it only require query time series length $N$ times 1-NN queries.

\begin{algorithm}[t]
    \caption{Defense Algorithm: $PatternHide$}\label{alg:defense}
 \begin{algorithmic}[1]
    \STATE \textbf{Input}: $T$, $\text{MPD}$, $\text{MPI}$, $L_{perm}$, $L_{attack}$
    \STATE \textbf{Output}: $\text{PMP}$ = \{$\text{MPD}$, $\text{MPI}$\}
    \STATE $\mathcal{C}$ = GetCIB($\text{MPI}$)
    \FOR{Each $C_i \in \mathcal{C}$ and $C_i>L_{perm}$}
    
    \\{\color{blue} /* Obtain Sensitive Intervals*/}
    \STATE $\mathcal{C}_{neighbor}=\{C |\quad |C.start - C_i.start|<L_{attack}\}$
       
    \STATE $S$ = concate($C_i\cup C_{neighbor}$)
       
    \STATE $\mathcal{S}$.add($S$)
    
    \ENDFOR
     \FOR{$S \in \mathcal{S}$}
     \STATE  $M=\{\}$
     \FOR{$idx$ in $[S.Start$, $ S.End]$}
     \STATE $L_{rnd}\sim U(0,L_{perm})$
        
     \IF{$\text{ConsecutiveIdxCount(idx)} > L_{rnd}$}
     \\{\color{blue} /* Compute Masked Matrix Profile*/}
          \STATE $M$.add(OverlappingIntervals($\text{MPI}(idx)$))
          \STATE $\text{MPD}'$, $\text{MPI}'$ = MaskMatrixProfile($T$, $M$)
          \\{\color{blue} /* Replace Original Matrix Profile*/}
          \STATE $\text{MPD}[idx:S.End]=\text{MPD}'[idx:S.End]$
          \STATE $\text{MPI}[idx:S.End]=\text{MPI}'[idx:S.End]$
     \ENDIF
         

      \ENDFOR
  \ENDFOR
  \\{\color{blue} /* Resolve any Symmetric Conflicts in Cycle Link Indexes by In-Place Update*/}
      \STATE $\text{MPD}$, $\text{MPI}$ = FakeCycleLink($\text{MPD}$, $\text{MPI}$)

    \STATE \textbf{return} $\text{MPD}$, $\text{MPI}$
  \end{algorithmic}
\end{algorithm}
\noindent\textbf{2) Resolve Conflicts in MPD} Every $\text{MPD}$ value is a distance and MPI may form a `cycle' (i.e., given two subsequences $S_{i}$ and $S_{j}$, $\text{MPI}(i) = j$ and $\text{MPI}(j) = i$. In this case, we need to enforce distance symmetry constraint $\text{MPD}(j) = \text{MPD}(i)$, otherwise a smart attacker might use this loophole to infer the modified locations. Therefore, we further generate fake symmetric distances (Line 25). Finally, the algorithm identifies any `cycles' in the MPI and checks if $\text{MPD}(i)=\text{MPD}(j)$ constraint is violates or not. If the distances are not equal, it adjusts $\text{MPD}$ value to be the minimum of the two. 

\subsection{Advantage of Proposed Algorithm}
Our defense algorithm is carefully designed and has two advantages attributed to the masked MP replacement. First, our algorithm plays a specific defense strategy to the attack method on protecting the location of large motif index, while keeping utility task performance in mind. The replaced values are still meaningful as similarity information can be seen as an approximation to the information recorded by actual MPD and MPI. Second, our strategy only perturb a small amount of index, so there is less information loss compared with the original MP. 

\vspace{-2mm}
\section{Experimental Evaluation}

In this section, we demonstrate that the proposed defense methods can successfully defend the proposed two attacks while maintaining the utility task performance on both real-world and synthetic data. Unless otherwise specified, the parameter $L_{perm}$ is set to $L_{perm}=L_{util}/4$.

\vspace{-2mm}
\subsection{Detecting Planted Motif while Protected Pattern Leakage}
\label{sec:8.1}
\vspace{-1mm}
We first evaluate the proposed defense method in motif discovery. Specifically, the utility task in the experiment is detecting motifs of length $L_{utility}$ while keeping the motif of length $L_{attack}$ protected. Following the previous planted motif evaluation experiment setting  \cite{Gao2018,gao2018exploring,mueen2013enumeration}, we test our Privacy-Aware MP in two different  scenarios: \\
\textbf{1) Independent Scenario}: In this scenario, sensitive patterns are independent of non-sensitive patterns. We randomly planted two independent motifs of lengths $L_{util}$ and $L_{attack}$, each one with two instances, into a long time series. \\
\textbf{2) Correlation Scenario}:  In this scenario, the location of sensitive pattern is correlated with non-sensitive pattern. We randomly planted a motif of length $L_{attack}$ of two instances, and a motif of length $L_{util}$ of three instances into the time series. Different from first experiment, two out of three instances of the short motif overlap with the long motif. Following the setting in large-scale planted motif experiment, the shape of motifs are randomly generated by using: $p=\sum_{i=1}^{5}A_i\sin(\alpha_i x+\beta_i)$,
with random parameters $A_i \in [0,10]$, $\alpha_i \in [-2,2]$ and $\beta_i \in [-\pi,\pi]$. Each instance of a motif,  $\pm$5\% random noise is added. In both experiments, the length of the random walk time series is 10,000. We test four different $L_{attack}=\{200,300,400,500\}$ while keeping the utility length $L_{util}=100$. All the experiments were repeated 50 times with different time series and motif shapes. 
\vspace{-3mm}
\subsection{Evaluation Criteria}
There are two evaluation criteria: average utility task performance and average attack success rate. Both criteria are evaluated based on motif detection rate.
The overlapping rate is measured by Jaccard similarity index: $J(pred,gt)=\frac{pred\cap gt}{pred \cup gt}$. If the overlapping rate of the detected location and ground truth is greater than 0.25 in both instances, we say the motif is detected. The average attack success rate is computed based on the success of locating motif of length $L_{attack}$ in the 50 experiments for each length. The utility task performance is measured by average motif discovery success rate of length $L_{util}$.
\vspace{-2mm}
\subsection{Baselines}
To the best of our knowledge, this is the first work that investigates the approach to prevent information leakage from deducing long pattern. None of existing approaches are designed for this problem. Therefore, we compare with two simple baselines: (1) \textbf{directly sharing MP} and (2) \textbf{adding noise in raw data}. For the second approach, we test three different variations, by adding small, medium, and large amounts of noise, respectively.
\vspace{-2mm}
\subsection{Attack Methods}
For MP sharing strategies, we evaluate the defense performance based on the success rate of two attack methods introduced in Sec. 6. For the approach that directly shares the matrix profile, we evaluate the performance by the success rate of directly detecting motif of $L_{attack}$ given the shared time series.
\vspace{-2mm}
\subsection{Vs. Sharing Original Matrix Profile}
We first compare the proposed PMP with the original MP. We apply both attacks described in Section 6 and report the attack success rate. Fig. 4(a)-(b) show the attack success rates in the non-overlap setting. According to the figure, PMP can significantly reduce the attack success rate. When the sensitive pattern length increases, the attack success rate on sharing MP decreases. This is because the correlation between MP of $L_{util}$ and $L_{attack}$ is much smaller when the length increases, making it harder for attackers to retrieve information. However, our attack methods still maintain up to 70\% success rate. Compared with sharing the original matrix profile, PMP maintains a stable low attack success rate (less than 0.15). Fig. 4(c)-(d) show the attack success rates in the overlap setting. Similar observation is found when we test the overlapping case. Sharing original matrix profile will lead to at least 0.7 success attack rate while the proposed approach can significantly reduce the chance of success in attack (up to 0.21). Moreover, the utility of PMP is shown in Table 1. From the table, it is about 0.88, which indicates our defense method can successfully defend the attacks while keeping the utility of the shorter segments. 

\begin{figure}[t] \centering
\subfigure[Entropy-based Attack \label{info_attack}]
  {\includegraphics[width=40mm]{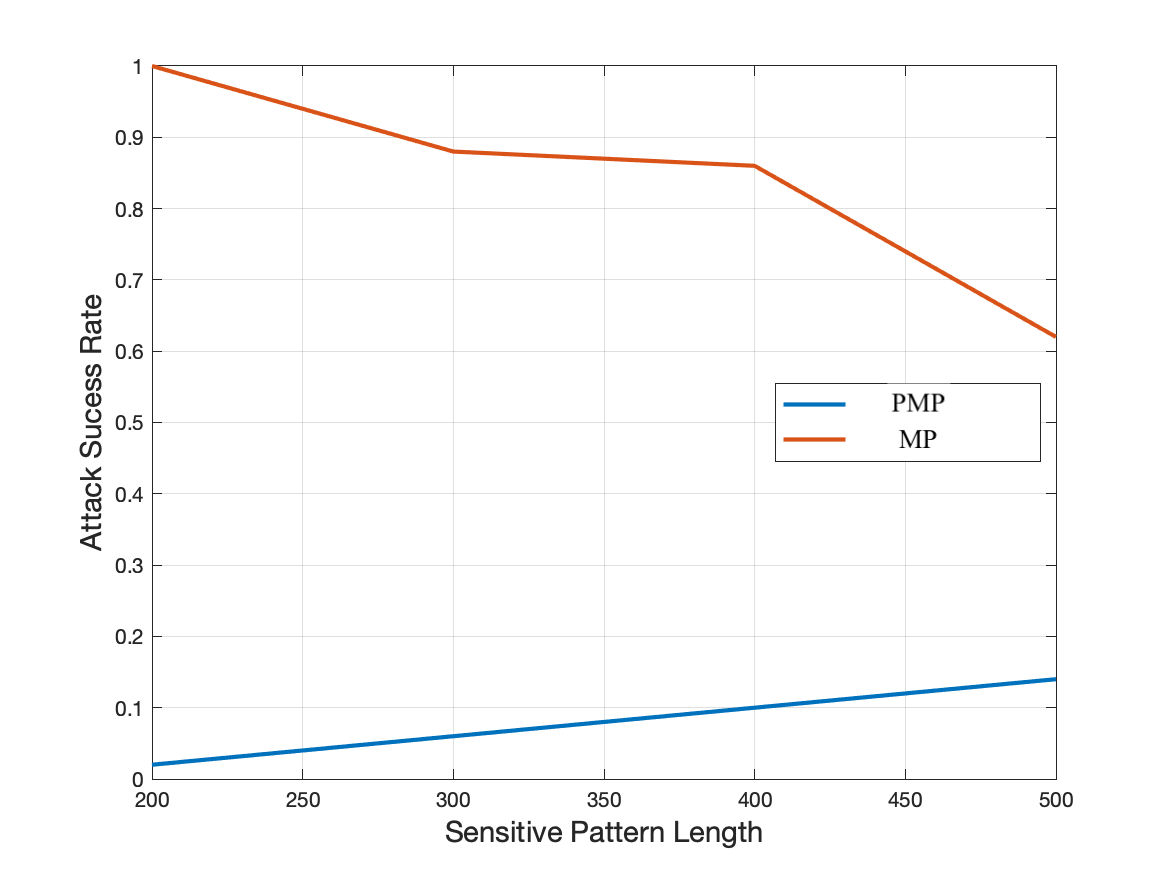}} 
\subfigure[Location-based Attack \label{info_attack2}]
  {\includegraphics[width=40mm]{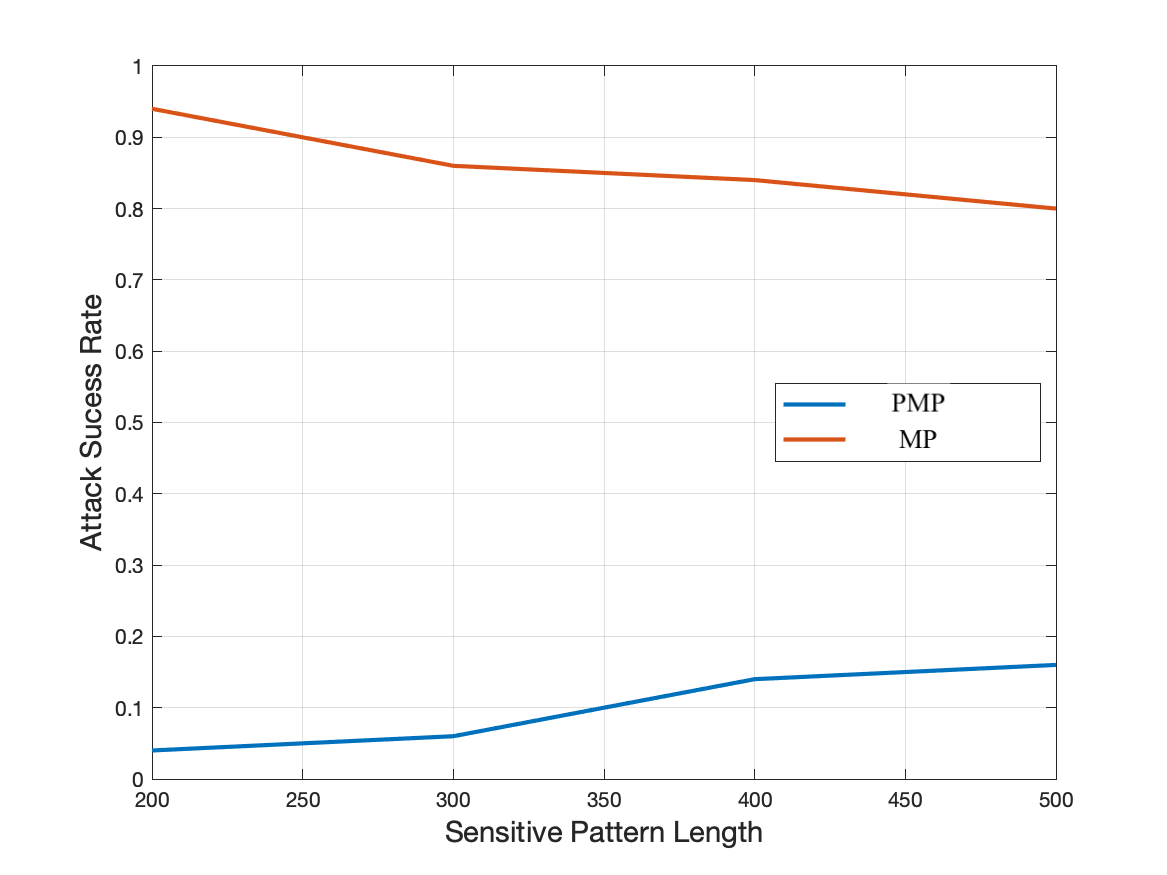}} 
\subfigure[Entropy-based Attack \label{info_attack3}]
  {\includegraphics[width=40mm]{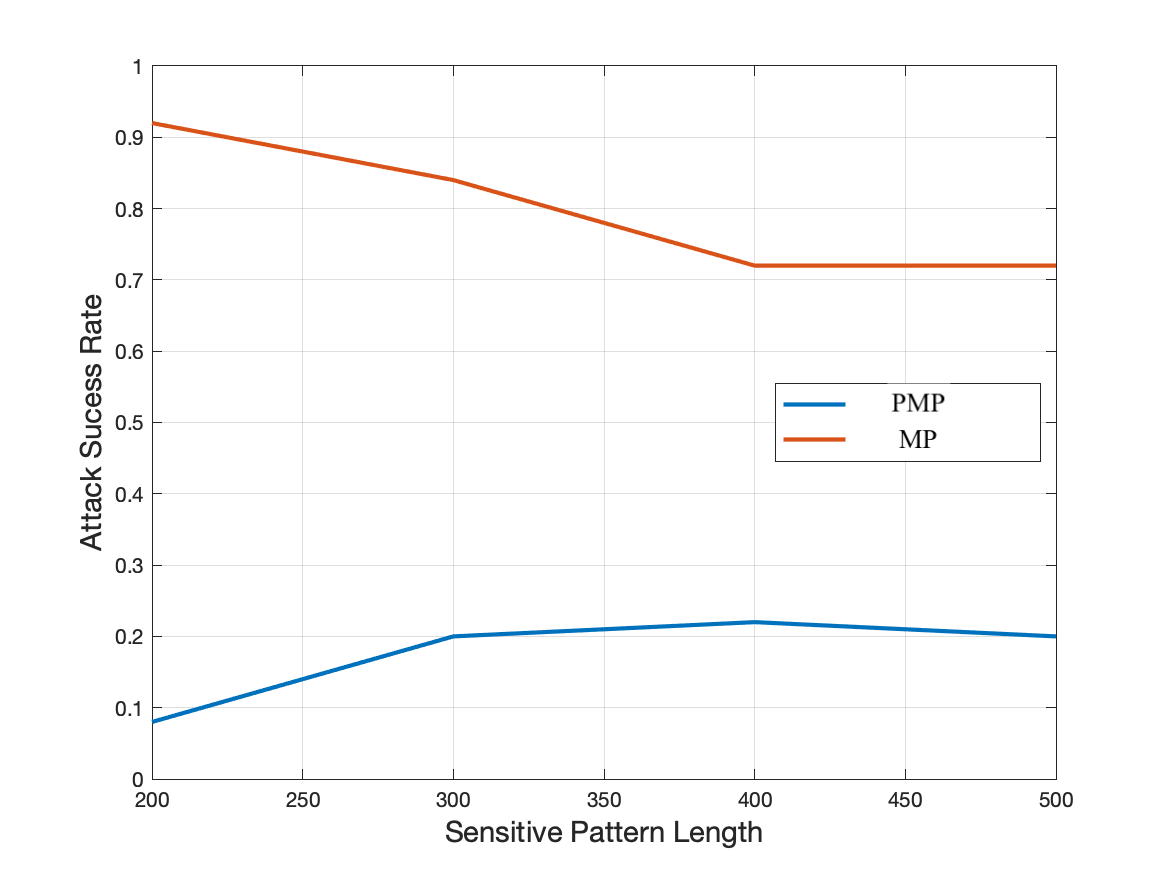}} 
\subfigure[Location-based Attack \label{info_attack4}]
  {\includegraphics[width=40mm]{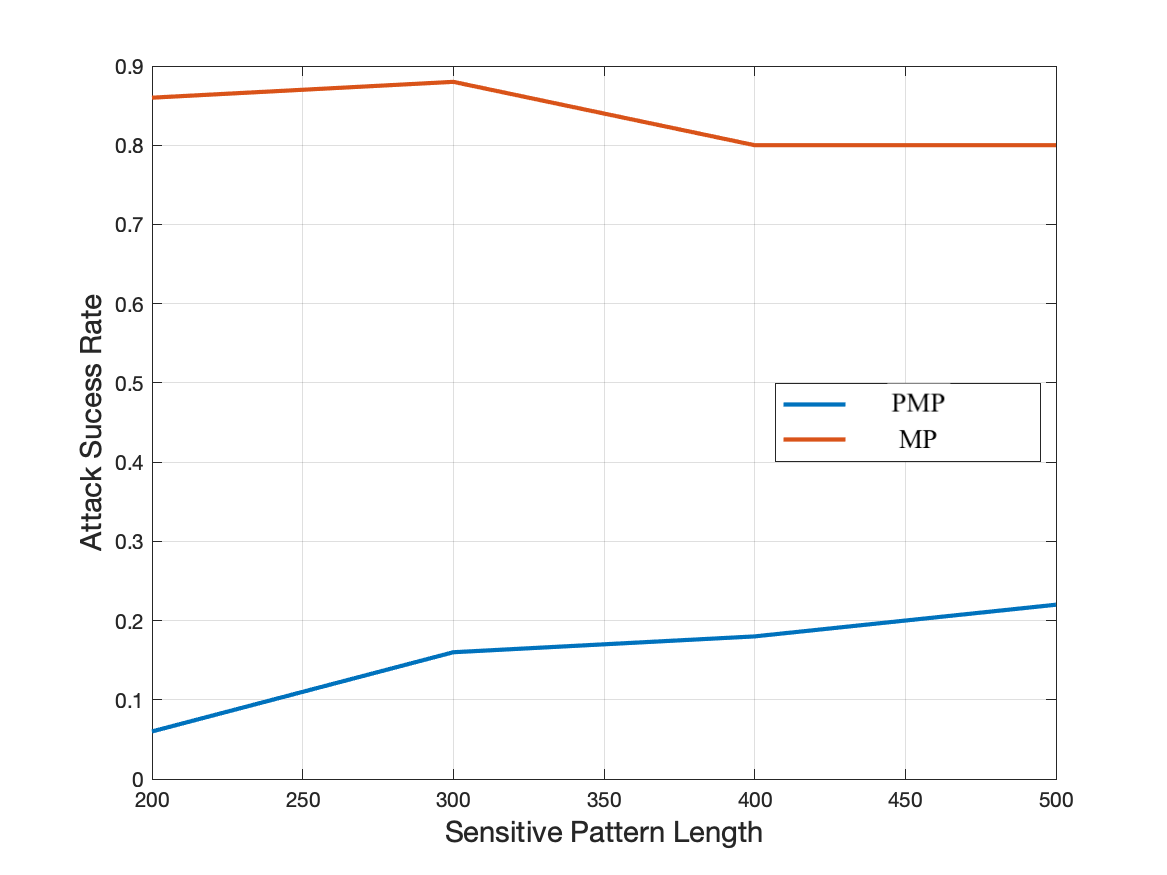}} 
 \vspace{-2mm} 
 \caption{Compared with Sharing MP under Different $L_{attack}$ (a-b) Independent Scenario, (c-d) Correlation Scenario.}
 \vspace{-3mm}
 \label{fig:exp1}
\end{figure}

\subsection{Vs. Perturbed Raw Series} \label{adding_noise}

We next test our proposed method with perturbed raw data. Specifically, we compare with sharing raw data after adding Gaussian noise with variance $\sigma^2 = \{0.1, 0.3, 0.5\}$, respectively. The attack success rates for all three cases are shown in Fig. 5(a)-(b) for both experiments. The utility task performance is shown in Table~1. 

In both experiments (Independent Scenario and Correlation Scenario as explained in Section 8.1), adding only small amounts of noise will result in high attack success rate and high utility of the data. When more noise is added, both attack success rate and the utility decrease. However, the utility decreases faster than the attack success rate. This is because small utility length pattern is much easier to be affected compared with long pattern. Moreover, compared with noise-adding approach, sharing PMP could maintain a very high level of utility (0.875 success rate on motif detection) while keeping the attack success rate close to the best defense performance of sharing perturbed time series. The experiment shows that the proposed approach outperforms the perturbed raw time series protocol. 

\begin{figure}[h]
\centering
 \subfigure[\label{att_nonoverlap}]
 {\includegraphics[width=42mm]{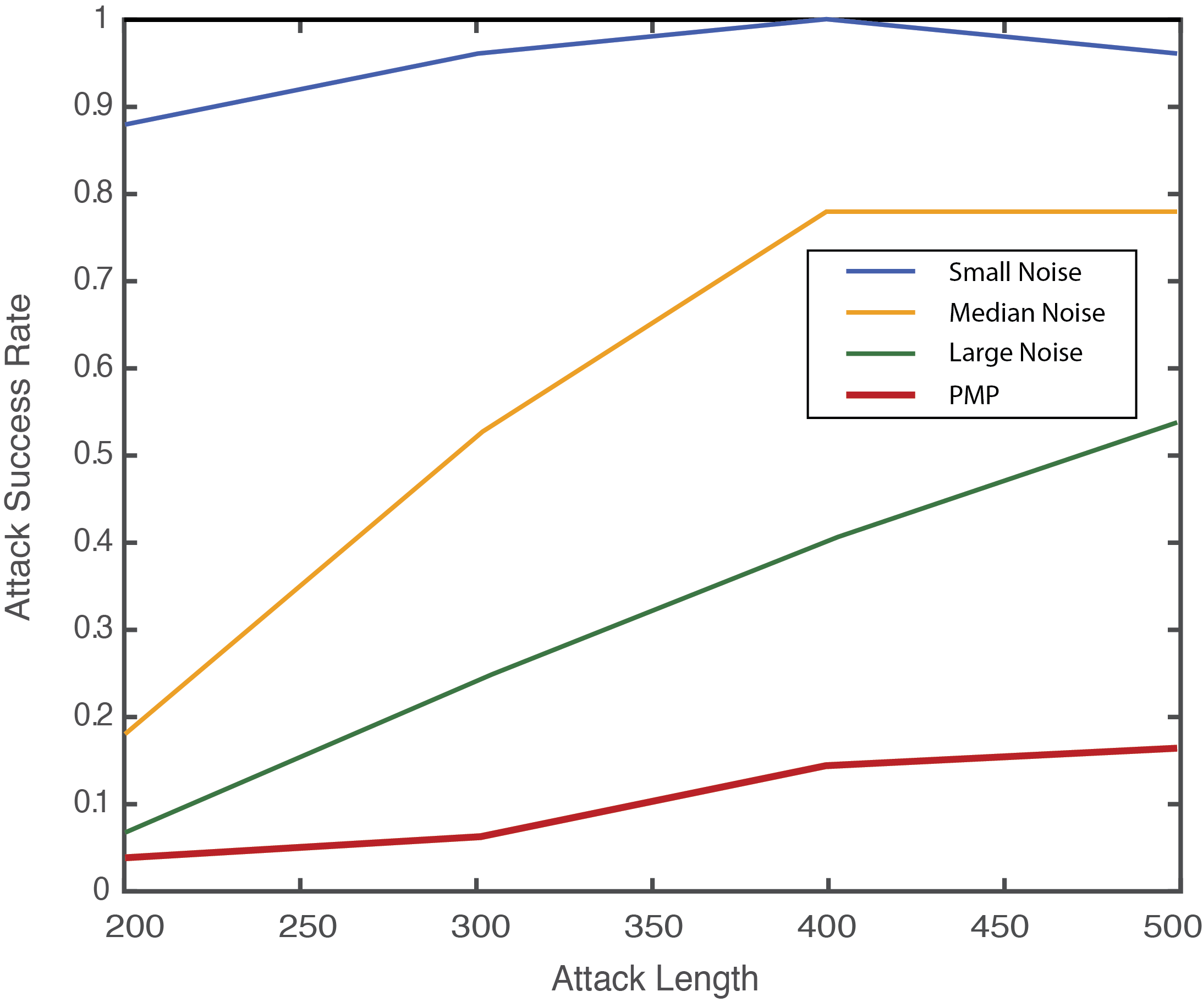}
 }
\subfigure[\label{att_overlap}]
  {\includegraphics[width=40mm]{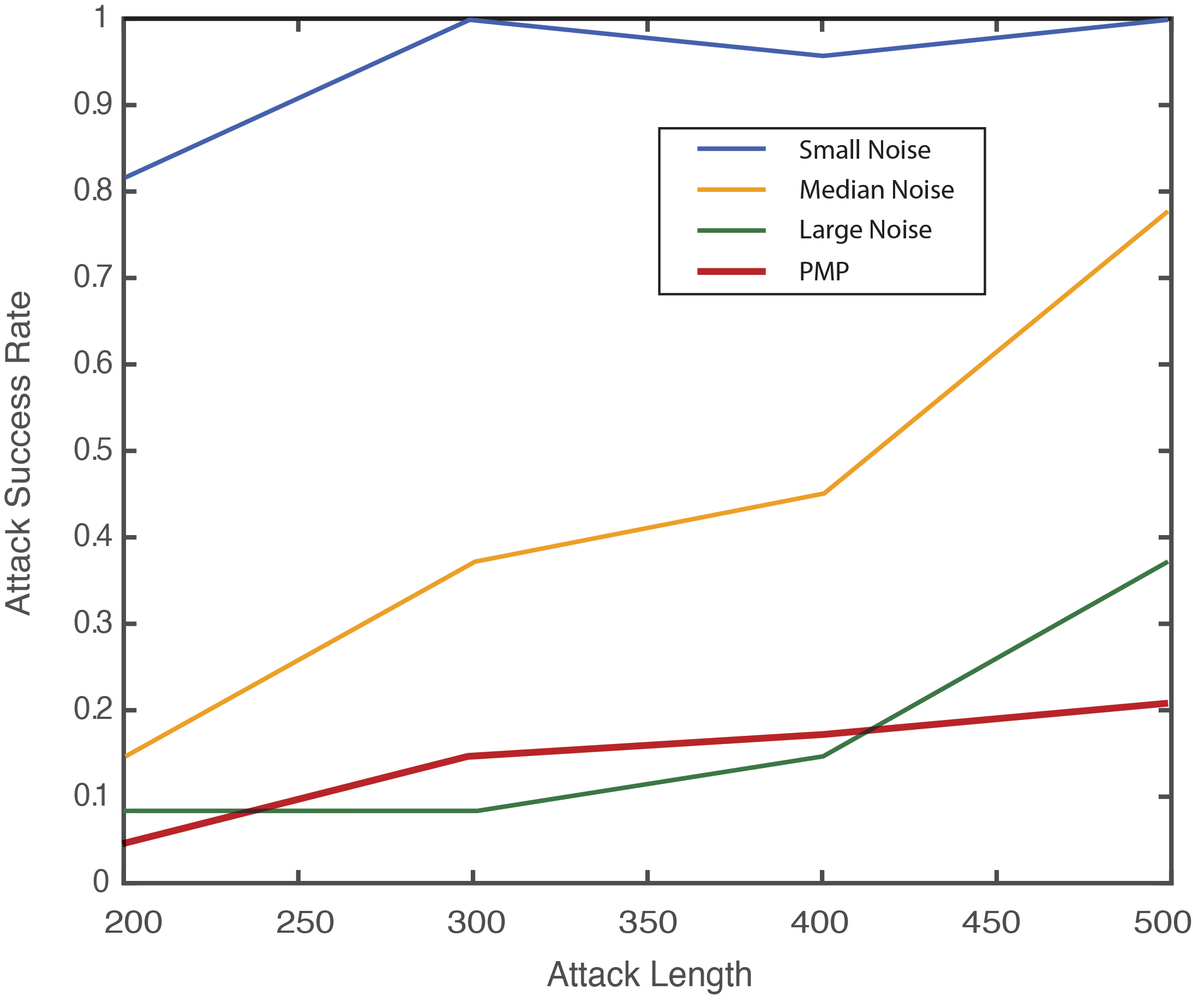}} 
 \caption{Compared with Sharing Perturbed Time Series \textbf{(a)} Independent Scenario \textbf{(b)} Correlation Scenario} \label{fig:exp3}
\end{figure}
\vspace{-3mm}
\begin{table}[h]
\centering
\caption{Shared PMP vs. Perturb Raw Time Series}
\vspace{2mm}
\scalebox{0.7}{
\begin{tabular}{|c|c|c|c|c|}
\hline
Setting \ Method & small $\sigma$ & median $\sigma$ & large $\sigma$ &  PMP\\
\hline\hline
Independent (utility) & 0.84       & 0.32         & 0.115        & \textbf{0.875}  \\    
Correlation (utility)  & 0.7       & 0.295       & 0.08        & \textbf{0.882}  \\   
\hline
\end{tabular}}
\end{table}

\begin{figure}
    \centering
    \includegraphics[width=80mm]{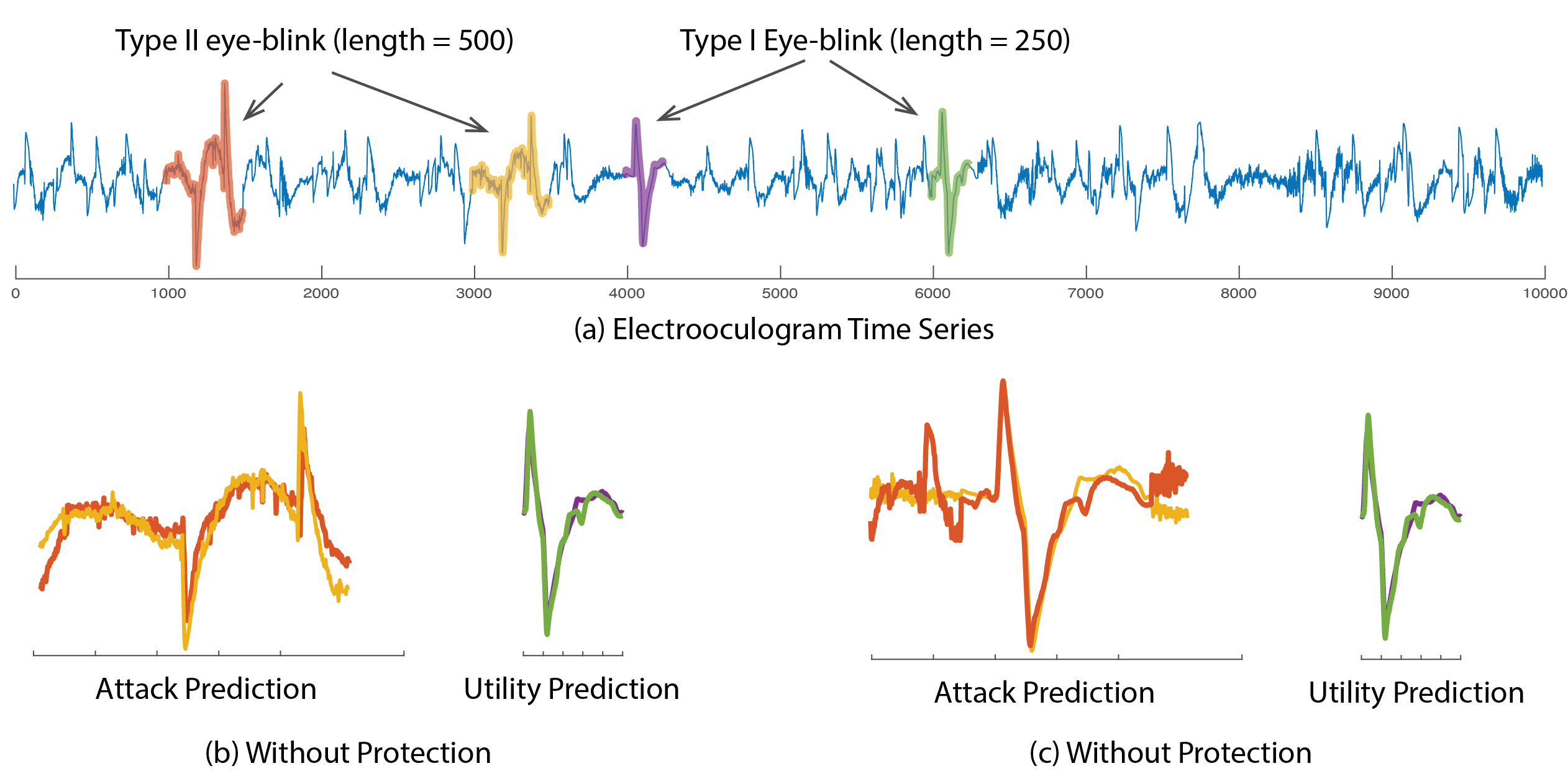}
    \caption{Protected MP successfully Prevent Type II Eye Blink Activity Leak}
    \label{fig:mp-to-data}
\end{figure}

\subsection{Motif Discovery on Electrooculogram Time Series}
In sleep quality study, Electrooculogram is a popular data type to study the sleep behavior of a subject~\cite{madrid2019matrix}. In this case study, we test our proposed method on preventing the leakage of long eye blinking activities from the Electrooculogram time Series. We utilized the EOG time series used by Madrid et al.~\cite{madrid2019matrix}. According to the authors, the time series is captured from a 66- year old healthy male recorded during a sleep study. The time series is shown in Fig. 6(a). Two types of motifs which correspond to two different types of eye blinking activities are highlighted in the time series. Without any perturbation, it is very easy for the attacker to retrieve the type II eye blinking pattern through the location/entropy based attack method. The detected pattern is shown in Fig. 9(b). After the perturbation, the attack algorithm fails to detect the long pattern. In fact, it locates the pattern somewhere overlapped with the small motif. Therefore, the proposed perturbed Matrix Profile protects the data from the attacker.

\vspace{-3mm}
\section{Conclusion}
In this paper, we consider a new type of privacy preserving problem in time series, i.e., preventing malicious inference on long shape-based patterns while preserving short segment information for the downstream task. To deal with the above problem, we introduced a shared Matrix Profile (MP) approach by utilizing the characteristics of MP as a stand-alone and versatile intermediate features. However, we illustrated that MP index sharing can still reveal the location of sensitive long pattern information based on two proposed attack methods. 
We further proposed a Privacy-Aware Matrix Profile (PMP) based on perturbing the index of matrix profile at sensitive pattern regions. 
We evaluated our proposed PMP sharing solution with several classic
defense methods through quantitative analysis and demonstrated the effectiveness of protecting sensitive information in
the real-world case study. This work will be a good first step that will hopefully inspire new interesting research for privacy-aware shape-based time series data mining methods.

\vspace{-4mm}
\bibliographystyle{abbrv}
\bibliography{TS-dp}
\end{document}